\documentclass{article}

\usepackage[margin=1in]{geometry}

\usepackage[utf8]{inputenc} 
\usepackage[T1]{fontenc}    
\usepackage{hyperref}       
\usepackage{url}            
\usepackage{booktabs}       
\usepackage{amsfonts}       
\usepackage{nicefrac}       
\usepackage{microtype}      
\usepackage{graphicx}
\usepackage{xcolor}
\usepackage[font=sf, labelfont={sf,bf}]{caption}

\usepackage{pgfplots}
\usepgfplotslibrary{groupplots}
\usetikzlibrary{patterns}

\usepackage{multirow}   
\selectcolormodel{cmyk}

\pgfplotsset{every axis/.append style={
                    xlabel={$x$},          
                    ylabel={$y$},          
                    label style={font=\sffamily\scriptsize},
                    tick label style={font=\sffamily\scriptsize}  
                    },
                    legend image code/.code={
                        \draw[mark repeat=1,mark phase=1]
                    }
                    }
\pgfplotsset{compat=newest}

\newcommand*{\eg}{{\em e.g.}}
\newcommand*{\ie}{{\em i.e.}}

\title{Posthoc Verification and the Fallibility of the Ground Truth}

\author{%
  Yifan Ding\quad Nicholas Botzer\quad Tim Weninger \\
  Department of Computer Science and Engineering\\
  University of Notre Dame\\
  Notre Dame, IN 46530 \\
  \texttt{\{yding4, nbotzer, tweninger\}@nd.edu} \\
}

\begin{document}

\date{}

\maketitle

\begin{abstract}
    Classifiers commonly make use of pre-annotated datasets, wherein a model is evaluated by pre-defined metrics on a held-out test set typically made of human-annotated labels. Metrics used in these evaluations are tied to the availability of well-defined ground truth labels, and these metrics typically do not allow for inexact matches. These noisy ground truth labels and strict evaluation metrics may compromise the validity and realism of evaluation results. In the present work, we discuss these concerns and conduct a systematic posthoc verification experiment on the entity linking (EL) task. Unlike traditional methodologies, which asks annotators to provide free-form annotations, we ask annotators to verify the correctness of  annotations after the fact (\ie, posthoc). Compared to pre-annotation evaluation, state-of-the-art EL models performed extremely well according to the posthoc evaluation methodology. Posthoc validation also permits the validation of the ground truth dataset. Surprisingly, we find predictions from EL models had a similar or higher verification rate than the ground truth. We conclude with a discussion on these findings and recommendations for future evaluations.


\end{abstract}

\section{Introduction}

Recent advances in machine learning are due to the wide availability of large labeled datasets. These datasets are typically curated from artifacts of digital human behavior that arise naturally, as in the case of click-through data or Wikipedia comments. But in many cases, these annotations are derived from crowd-sourced tasks where annotations are provided by amateur workers~\cite{maas2011learning, bowman2015large, richardson2013mctest, bowman2015large,rajpurkar2016squad, trischler2016newsqa, geva2019we}. Because they rely on subjective human judgement, the processes used to construct these datasets are inherently error-prone~\cite{sambasivan2021everyone}. Strategies to decrease the error rate have been developed~\cite{hynes2017data}, but errors persist. Recent work has focused on training models using noisy, subjective, human-labeled labels~\cite{song2020learning}, and many have concluded that deep learning systems are robust to noisy labels -- so long as the training data is not too small and not too noisy~\cite{rolnick2017deep}.

Understanding and mitigating noise in benchmark datasets remains a major focus of research~\cite{oortwijn2021interrater}. For example, in natural language processing (NLP) several approaches have been developed that are robust to noise~\cite{belinkov2018synthetic}, logically consistent~\cite{ribeiro2019red}, explainable~\cite{lundberg2017unified}, or fair~\cite{prabhakaran2019perturbation}. Oftentimes only a handful of annotators produce most of the dataset annotations; this is known to result in annotator bias that can severely influence model performance and generalizability~\cite{levy2015improving, schwartz2017story, gururangan2018annotation, glockner2018breaking, poliak2018hypothesis, tsuchiya2018performance, aharoni2018split, paun2018comparing, geva2019we}. And sometimes human annotations are just wrong~\cite{rosales2019nifify}. 

These efforts have lead to the development of machine learning systems that are more robust to noisy labels; unfortunately, annotations that appear in the held-out test set are typically considered to be correct even though they likely have the same noise as the training set~\cite{rajpurkar2018know, northcutt2021pervasive}. Pervasive annotation errors in benchmark datasets could undermine the framework by which we validate performance and measure progress in machine learning~\cite{northcutt2021pervasive}.

\paragraph{What are label errors? Where do they come from? How do we know?}
The general machine learning pipeline starts with a dataset (a collection of documents, images, medical records, etc.). When labels are not inherent to the data, they must be annotated -- usually by humans. A label error occurs when an annotator provides a label that is ``incorrect.'' But this raises an interesting question: who gets to decide that some annotation is incorrect? One solution is to ask $k$ annotators and combine their labels somehow (\eg, majority vote, probability distribution). 

Subjectivity comes into play here. Given identical instructions and identical items, some annotators may focus on different attributes of the item or have a different interpretation of the labeling criteria. Understanding and modelling label uncertainty remains a compelling challenge in evaluating machine learning systems~\cite{sommerauer2020would}.

\paragraph{Soft errors and subjectivity.} Tasks that require free-form, soft, or multi-class annotations present another dimension to this challenge. For example, when providing labels for an object detection task the annotator is typically asked to draw a box around some object (\eg, identify the bicycle in this photo). These annotations are then used to train an object detection model, which will produce labels that are unlikely to exactly match the ground truth annotations. Image processing researchers have therefore developed ways to cope with inexact matches involving geometric assessments of overlap and a threshold that defines ``close enough'' for precision, recall, and other hard metrics~\cite{wang2020image}.  But the concept of ``close enough'' in a geometric sense may not match what an annotator or end-user might consider to be ``close enough'' in a semantic sense.

\begin{figure}
    \centering
    \includegraphics[width=.85\textwidth]{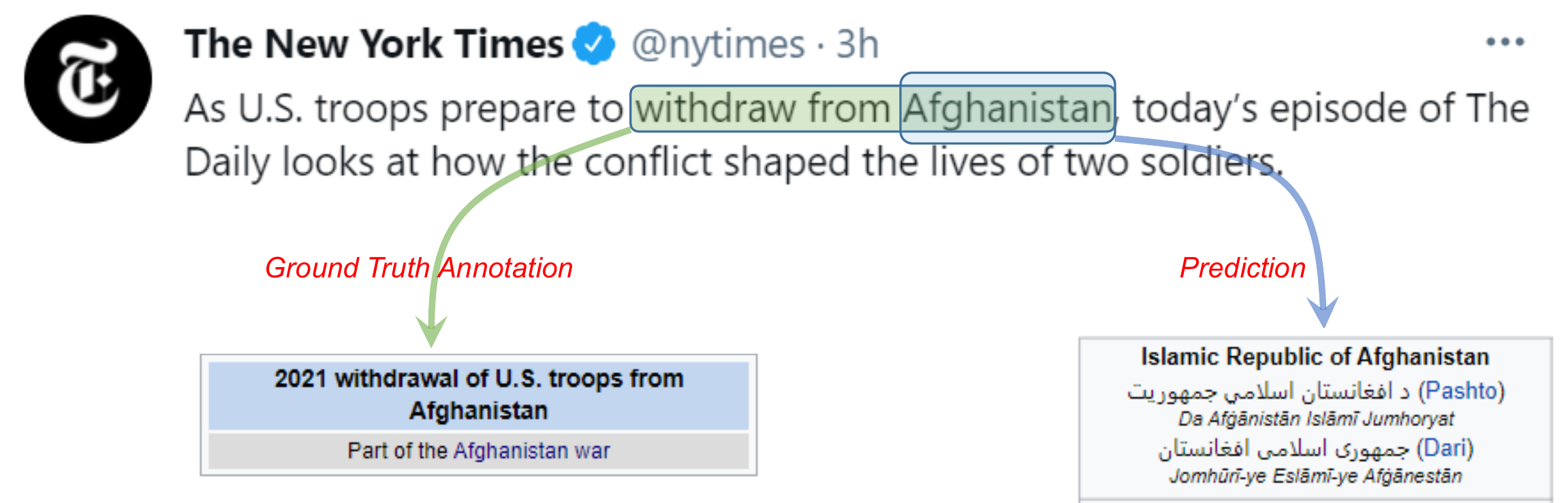}
    \caption{Example Entity Linking task where the pre-annotated ground truth mention, represented by bounding boxes around text, and link, represented by arrows drawn to Wikipedia entries, is different from the predicted label. Standard evaluation regimes count this as a completely incorrect prediction despite being a reasonable label.}
    \label{fig:posthoc_example}
\end{figure}

Likewise, natural language processing tasks like named entity recognition (NER) and entity linking (EL) rely heavily on datasets comprised of free-form human annotations. As in object detection, NER and EL tasks are typically evaluated against a held out portion of the already-annotated dataset. Unfortunately, NER and EL tasks produce labels that are not easily verified as ``close enough''~\cite{ribeiro2020beyond}. Instead, like the example in Fig.~\ref{fig:posthoc_example}, most NER and EL evaluation metrics require exact matches against free-form annotations~\cite{sevgili2020neural, goel2021robustness}. This strict evaluation methodology may unreasonably count labels that are ``close enough'' as incorrect. 

A similar problem is encountered in machine translation and language summarization where a translated and summarized sentences are not easily matched to pre-annotated labels. Instead, these tasks use Direct Assessment (DA) to rank and compare model predictions~\cite{graham2014machine, graham2013continuous}. But, as machine learning systems advance, ranking equally-valid predictions becomes difficult. Instead, if we simply ask an annotator \textit{posthoc} to verify whether or not a predicted annotation was correct or useful, then the task becomes easier to annotate while still accomplishing the goal. 

\paragraph{Correct versus Verifiable.} Producing a \textit{verifiable} answer is not the same as producing the \textit{correct} answer. This distinction is critical. Asking a machine learning system to independently provide the same label as an annotator is a wildly different task than asking an annotator to verify the output of a predictor (\textit{posthoc verification}). 

Unfortunately the prevailing test and evaluation regime requires predictors to exactly match noisy, free-form, and subjective human annotations. This paradigm represents a mismatch that, if left unaddressed, threatens to undermine future progress in machine learning.

\paragraph{Main Contributions.} In the present work we show that these distinctions matter and that they present significant consequences for how we determine the state-of-the-art in machine learning systems. Specifically, we ask three research questions: 

\begin{enumerate}
    \item[RQ1] When evaluating machine learning systems, how do the results of posthoc verification compare to standard validation against a pre-annotated ground truth? How does this impact validation of the state of the art?

    \item[RQ2] How much agreement is there among posthoc annotators? How does it compare to the annotation agreement on the ground truth datasets?

    \item[RQ3] What do verification rates say about the ground truth data? Do some datasets \textit{perform} better or worse than others?
\end{enumerate}

Using a large case study of eight popular EL datasets and two state-of-the-art EL models we are able to report some surprising findings: First, state-of-the-art EL models generally had a \textit{higher} verification rate than the ground truth datasets. Second, there was substantial disagreement among annotators as to what constitutes a label that is ``close enough''. Third, a large proportion (between 10\%-70\% depending on the dataset) of verified entities were missing from the ground truth dataset. 


In the remainder of this paper we will highlight some related perspectives machine learning evaluation and then describe our posthoc verification methodology. Next, each of the research questions will be addressed. Finally, we conclude the paper with a short discussion including our recommendations for future work.








\section{Posthoc Verification}

In a time where hyper-tuned models measure improvement in tenths of a percent, many are beginning to recognize the importance of benchmark datasets. This realization has inspired many to wonder, given the limitations of the benchmarks, if state-of-the-art systems have reached their maximum achievable values~\cite{devlin2019bert, radford2019language}. These findings have led to the development of iterative benchmarks that collect adversarial data to fool some models~\cite{kiela2021dynabench}. But others argue that this iterative/adversarial model creates a ``counterproductive incentive to be different without being better''~\cite{bowman2021will}. Yet, in the search for better evaluation methodologies, the annotation question has remained the same, \ie, the annotator is asked to provide a label (or labels) for an item (or set of items). 

On the contrary, \textit{posthoc verification} supplies prediction to the annotator, who is then asked to verify it accuracy or not. By supplying a prediction to the annotator, this methodology asks for similar information, but in a critically different way. 

In the broader literature on survey methodology and human behavior the difference between pre-annotation and posthoc verification is probably best captured by the default effect. The default effect is the tendency for an agent to generally accept the options that are presented to them~\cite{park2000choosing}. In this frame, the choice of rejecting a pre-defined label is a more deliberate action than inputting a free-form annotation from scratch. 

There is much to learn from this distinction. To that end, the present work explores the differences between pre-annotation and posthoc verification methodologies by focusing on the entity linking (EL) task. Despite this focus, we expect that our results generalize to any prediction task that makes use of free-form, subjective, or many-labeled benchmark datasets.

\paragraph{Datasets.}

\begin{table}
\centering
{\footnotesize\addtolength{\tabcolsep}{-1pt}
\caption{Statistics of the entity linking datasets and annotations.}
{\renewcommand{\arraystretch}{1.1}
\vspace{.1cm}
    \begin{tabular}{ll | l | lll  | lll | lll}
\toprule
& 
\multirow{2}{*}{\textbf{Datasets}} &
\multirow{2}{*}{\textbf{Docs}} &
\multicolumn{3}{c|}{\textbf{Annotations}} &
\multicolumn{3}{c|}{\textbf{Tasks}} &
\multicolumn{3}{c}{\textbf{Verified Annotations}} \\

&&&
\textbf{GT} & \textbf{E2E} & \textbf{REL} & \textbf{GT} & \textbf{E2E} & \textbf{REL} & \textbf{GT} & \textbf{E2E} & \textbf{REL} \\ \midrule
\multirow{3}{*}{\rotatebox[]{90}{\textbf{AIDA}}} 
 & \textbf{AIDA-train}         & 946 & 18541* & 18301 & 21204 & 2801 & 2802 & 2913 & 18511 & 18274 & 21172 \\
 & \textbf{AIDA-A}         & 216 & 4791 & 4758 & 5443 & 713 & 715 & 725 & 4787 & 4754 & 5439 \\
 & \textbf{AIDA-B}         & 231 & 4485 & 4375 & 5086 & 636 & 646 & 654 & 4480 & 4370 & 5079 \\
 \midrule
 \multirow{5}{*}{\rotatebox[]{90}{\textbf{WNED}}} 
 & \textbf{ACE2004}       & 57* & 257 & 1355 & 1675 & 114 & 318 & 334 & 256 & 1352 & 1672 \\
 & \textbf{AQUAINT}       & 50 & 727 & 810 & 925 & 175 & 170 & 179 & 727 & 810 & 925 \\
 & \textbf{CLUEWEB}       & 320 & 11154 & 12273 & 23114 & 3526 & 3678 & 4944 & 11139 & 12247 & 23056 \\
 & \textbf{MSNBC}         & 20 & 656 & 629 & 756 & 164 & 163 & 171 & 656 & 629 & 756 \\
 & \textbf{WIKIPEDIA}     & 345* & 6793* & 8141 & 11184 & 1348 & 1578 & 1638 & 6786 & 8136 & 11177 \\
 
\bottomrule
\end{tabular}
}
\label{tab:basic_statistics}
}
\end{table}

The goal of EL is to identify words or phrases that represent real-world entities and match that phrase to a listing in some knowledge base. Like most classification systems, EL models are typically trained and tested on large pre-annotated benchmark datasets. In the present work we identified eight popular benchmark datasets that are widely used throughout the EL and broader NLP communities. These datasets are listed in Table~\ref{tab:basic_statistics}. They include three datasets from the AIDA benchmark from a 2003 CoNLL competition~\cite{hoffart2011robust} and five datasets from the (random) Walking Named Entity Disambiguation (WNED) benchmarks curated by Guo and Barbosa~\cite{guo2014robust}. Each dataset contains many individual documents. This ground truth (GT) is derived from human annotators. Specifically, the AIDA benchmark was annotated by two students (with disagreements resolved by the PI)~\cite{hoffart2011robust}, and the WNED benchmark is annotated through various means. Entries marked with * indicate statistics that are different from those reported in related work~\cite{ganea2017deep} because related works filter out documents that do not have any annotations that appear their pre-defined dictionary. We do not perform this filtering.

\paragraph{Models}
In order to better understand the effect of pre-annotated benchmarks on machine learning systems, it is necessary to test a handful of state-of-the-art EL systems. Specifically, we chose:

\begin{enumerate}
    \item[E2E] The end-to-end entity linking model, which generates and selects span candidates associated with entity candidates. E2E is a word-level model that utilizes word and entity embeddings to compute span-level context scores. Word and entity embeddings are trained on Wikipedia, and the final model is trained and validated using AIDA-train and AIDA-A respectively~\cite{kolitsas2018end}.
    \item[REL] The Radboud Entity Linker combines the Flair~\cite{akbik2018contextual} NER system with the mulrel-nel~\cite{le2018improving} entity disambiguation system to create a holistic EL pipeline~\cite{van2020rel}. Mulrel-nel is trained on AIDA-train.
\end{enumerate}

In addition to these popular EL systems, our methodology permits the evaluation of the GT as if it were a competing model. The relative performance of E2E and REL can then compared with the GT to better understand the performance of the annotations.

\paragraph{Pre-Annotation Metrics}
Like many NLP tasks, EL evaluates model performance using standard macro and micro averaging of precision, recall, and F1 scores. These metrics typically require that predictions, \ie, the detected mentions and associated entity-links, exactly match the ground truth. Another evaluation methodology relaxes the exact-matching requirement and allows overlapping mentions (like the example from Fig.~\ref{fig:posthoc_example}) to be counted as correct prediction if they are linked to the same exact entity in the knowledge base (which the example from Fig.~\ref{fig:posthoc_example} does not)~\cite{rosales2019fine}.

Recently, the GERBIL tool has been created to give researchers an easy to use API to test EL models against many of these pre-annotated datasets~\cite{roder2018gerbil}. As we might expect, many users have found that the GERBIL benchmarks primarily feature generic and prominent entities~\cite{van2016evaluating}.

\subsection{Data collection}
\label{sec:collection}
We have previously argued that these evaluation metrics may not faithfully simulate \textit{in vivo} performance because (1) the ground truth annotations are noisy and subjective, and (2) exact matching is too strict. We test this argument by collecting posthoc verifications of the three models, including the pre-annotated GT, over the eight datasets.

\begin{figure}
    \centering
    \includegraphics[width=\textwidth]{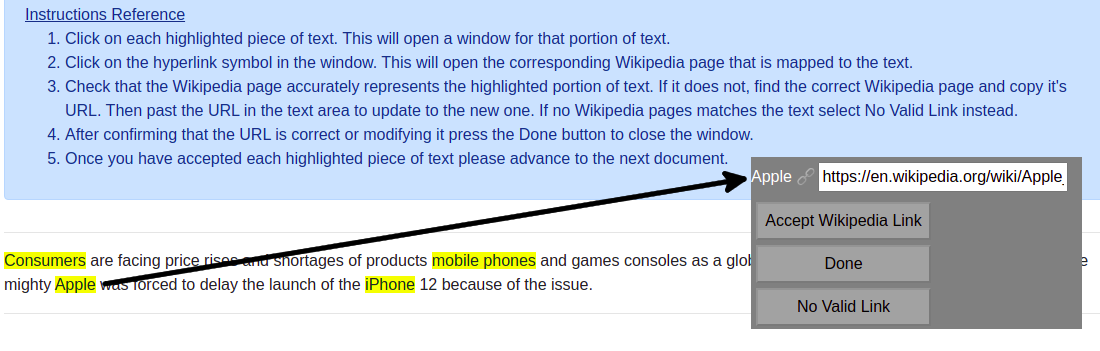}
    \caption{Example posthoc verification task. A worker is asked to click on each highlighted entity mention, which displays an entity-link. The worker may then verify, edit, or remove the link.}
    \label{fig:app_example}
\end{figure}

We created a simple verification system and used Amazon Mechanical Turk to solicit workers. For each document and model, we asked a single worker to verify all present entity annotations (\ie, an entity mention and its linked entity). Large documents of more than 300 characters were broken into smaller tasks so as to not overwhelm the worker. An example task with four annotations is illustrated in Fig.~\ref{fig:app_example}; where workers are asked to click on each annotation (highlighted in yellow), open up the corresponding Wikipedia page, and select one of three possible actions in the window:

\begin{enumerate}
    \item[Verify] The annotator determines that the current annotation (both mention and Wikipedia link) is appropriate.
    
    \item[Modify] The annotator determines that the Wikipedia link is incorrect. In this case, they are asked to search and select a more appropriate Wikipedia link, use it to replace the existing link, and then accept the new annotation.
    
    \item[Remove] The annotator determines that the current mention (highlighted text) is not a linkable entity. In this case, they remove the link from the mention.
\end{enumerate}

We made a deliberate decision to not permit new annotation of missing entity mentions. That is, if the model did not label an entity, then there is no opportunity for the worker to add a new label. This design decision kept the worker focused on the verification task, but possibly limits the coverage of the verified dataset. We provide further comments on this decision in the Results section.

Each annotator is assigned to 20 tasks including one control task with three control annotations. We only accept and collect annotations from workers that passed the control task. We paid each worker 3 USD for each HIT. We estimate a average hourly rate of about 9 USD; and paid a total of 6,520 USD. From these, we received 167,432 annotations. The breakdown of tasks, annotations shown to workers, and verified annotations are listed in Table~\ref{tab:basic_statistics} for each dataset and model. 

Prior to launch, this experiment was reviewed and approved by an impaneled ethics review board at the University of Notre Dame. The full list of instructions and link to an example video is included in the Supplementary Material. The source code, raw results, and evaluation scripts are publicly available via the MIT license at \url{https://github.com/yifding/e2e_EL_evaluate}

\section{Methodology and Metrics}

Using the posthoc verification dataset, we can begin to answer the three research questions. But before we directly address the specific questions, we first investigate the collected dataset and its relationship to the typical evaluation regimes. Then, in order to distinguish between traditional, pre-annotated evaluation metrics like precision and recall, we define (or rather redefine) similar metrics for the post-hoc verification scenario before discussing the posthoc verification results.

\subsection{The Pre-Annotation Evaluation Regime} 

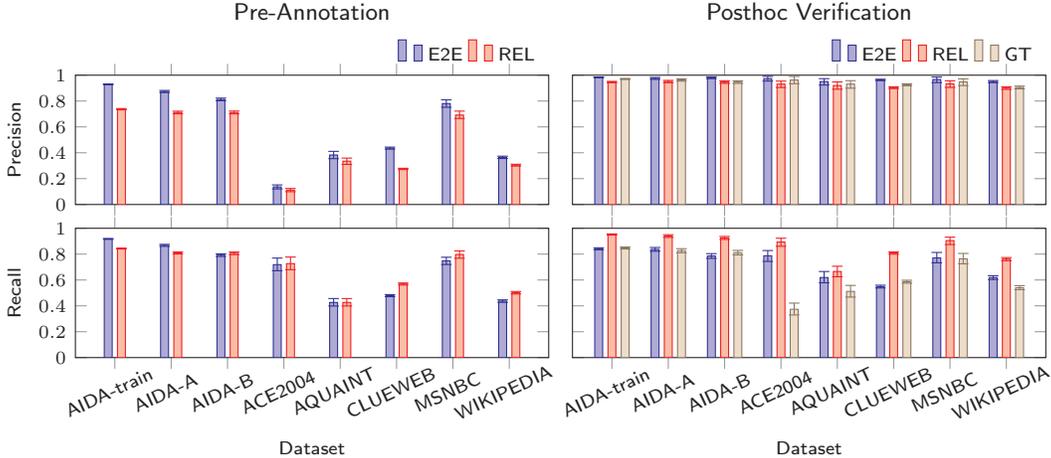
\begin{figure}
    \centering
    \pgfplotstableread{
metric	e2eprec	e2epreclow	e2eprechigh	e2erecall	e2erecalllow	e2erecallhigh	e2ef	relprec	relpreclow	relprechigh	relrecall	relrecalllow	relrecallhigh	relf
AIDA-train	0.9295173471	0.002955018058	0.003064463172	0.9176165523	0.003457403706	0.003241315974	0.9235286122	0.7372000756	0.005148309087	0.00477045154	0.8431743288	0.004267732699	0.004375776565	0.786634075
AIDA-A	0.8737904922	0.008203618006	0.008203618006	0.8677668686	0.007520367662	0.007938165866	0.8707682633	0.7117117117	0.0102960103	0.009744438316	0.8086484228	0.009400459578	0.009400459578	0.7570897712
AIDA-B	0.8123569794	0.009839816934	0.01029748284	0.7924107143	0.01004464286	0.009598214286	0.802259887	0.7115573932	0.01063201418	0.01102579248	0.8066964286	0.009598214286	0.009375	0.7561460404
ACE2004	0.1360946746	0.01553254438	0.01553254438	0.71875	0.046875	0.05078125	0.2288557214	0.1112440191	0.01196172249	0.01255980861	0.7265625	0.046875	0.046875	0.1929460581
AQUAINT	0.3827160494	0.02839506173	0.02839506173	0.4264099037	0.02751031637	0.03026134801	0.4033832141	0.3351351351	0.02486486486	0.02378378378	0.4264099037	0.03026134801	0.03026134801	0.3753026634
CLUEWEB	0.4357801911	0.007593696415	0.007920307014	0.4791273903	0.008079719903	0.007900170572	0.4564269221	0.2749826509	0.004857737682	0.005204718945	0.5691713798	0.007451297244	0.00754107191	0.3708144466
MSNBC	0.7806041335	0.02861685215	0.02861685215	0.7484756098	0.02896341463	0.02743902439	0.7642023346	0.6917989418	0.02645502646	0.03042328042	0.7972560976	0.02591463415	0.02591463415	0.7407932011
WIKIPEDIA	0.3644296952	0.00860373648	0.009341199607	0.4369289714	0.009578544061	0.009578544061	0.3973998124	0.3044645254	0.007157555695	0.00679967791	0.5014736222	0.009578544061	0.01031535514	0.3788899404
}{\prehoc}

\pgfplotstableread{
metric	e2eprec	e2epreclow	e2eprechigh	e2erecall	e2erecalllow	e2erecallhigh	e2ef	relprec	relpreclow	relprechigh	relrecall	relrecalllow	relrecallhigh	relf	gtprec	gtpreclow	gtprechigh	gtrecall	gtrecalllow	gtrecallhigh	gtf
AIDA-train	0.9832973079	0.002947533897	0.002947533897	0.8414326551	0.008071296452	0.007734992433	0.9068503081	0.9469899666	0.005016722408	0.005016722408	0.9522448293	0.004708256264	0.004371952245	0.9496101283	0.9708999807	0.003854307188	0.004047022548	0.8471498234	0.007398688414	0.007734992433	0.9048132184
AIDA-A	0.9752906977	0.00726744186	0.00726744186	0.8366583541	0.01433915212	0.01496259352	0.9006711409	0.9514195584	0.009463722397	0.008201892744	0.9401496259	0.01059850374	0.009975062344	0.9457510191	0.962962963	0.00871459695	0.007988380537	0.8266832918	0.01620947631	0.01558603491	0.8896343509
AIDA-B	0.9802575107	0.00686695279	0.00686695279	0.7859600826	0.01858224363	0.01789401239	0.872421696	0.9471830986	0.00985915493	0.00985915493	0.9256710255	0.01169993118	0.01101169993	0.9363035155	0.9469453376	0.01045016077	0.009646302251	0.8107364074	0.0165175499	0.01789401239	0.8735632184
ACE2004	0.9728506787	0.01809954751	0.01809954751	0.7875457875	0.04395604396	0.04029304029	0.8704453441	0.9312977099	0.02671755725	0.02290076336	0.8937728938	0.03296703297	0.0293040293	0.9121495327	0.9622641509	0.02830188679	0.02830188679	0.3736263736	0.04395604396	0.04761904762	0.5382585752
AQUAINT	0.9508928571	0.02678571429	0.02232142857	0.6191860465	0.04069767442	0.04651162791	0.75	0.9196787149	0.0281124498	0.0281124498	0.6656976744	0.04069767442	0.04069767442	0.7723440135	0.9312169312	0.03174603175	0.02645502646	0.511627907	0.04360465116	0.04651162791	0.660412758
CLUEWEB	0.9632653061	0.005337519623	0.005337519623	0.5484447622	0.01018948874	0.01108330354	0.6989406538	0.9036770584	0.007194244604	0.006594724221	0.8083661065	0.008759385055	0.008580622095	0.8533685601	0.9258110014	0.007334273625	0.007334273625	0.5867000358	0.01108330354	0.0112620665	0.7182405077
MSNBC	0.966507177	0.02392344498	0.01913875598	0.7709923664	0.03816793893	0.04198473282	0.8577494692	0.9330708661	0.02755905512	0.02362204724	0.9045801527	0.03053435115	0.02671755725	0.9186046512	0.9478672986	0.02843601896	0.02369668246	0.7633587786	0.03816793893	0.04198473282	0.8456659619
WIKIPEDIA	0.9502308876	0.008209338122	0.008209338122	0.6183639399	0.01469115192	0.01469115192	0.7491909385	0.9001972387	0.009861932939	0.009467455621	0.7619365609	0.01268781302	0.01168614357	0.8253164557	0.9049217002	0.01062639821	0.01062639821	0.5402337229	0.01502504174	0.01602671119	0.6765628267
}{\posthoc}

\begin{tikzpicture}
\sffamily
\begin{groupplot}[
group style={
        group name=my plots,
        group size=2 by 2,
        xlabels at=edge bottom,
        ylabels at=edge left,
        xticklabels at=edge bottom,
        yticklabels at=edge left,
        horizontal sep=9pt,
        vertical sep=9pt,
    },
ybar=2pt,
/pgf/bar width=3pt,
legend style={font=\scriptsize,
	nodes={scale=1, transform shape},
	at={(1,1)},
	anchor=south east,
	draw=none, fill=none},
legend columns=-1,
legend cell align={left},
width = 3.1in, 
height = 1.3in,
ymin=0.0, ymax=1,
ylabel = {},
xlabel = {Dataset},
symbolic x coords={
AIDA-train,
AIDA-A,
AIDA-B,
ACE2004,
AQUAINT,
CLUEWEB,
MSNBC,
WIKIPEDIA
},
xtick=data,
title style={yshift=.4cm,font=\small},
x tick label style={rotate=25,anchor=center,yshift=-0.25cm,xshift=-0.2cm},
]

\nextgroupplot[title={Pre-Annotation}, ylabel={Precision}]
\addplot+[error bars/.cd, y dir = both, y explicit] table[x=metric, y=e2eprec, y error minus=e2epreclow, y error plus=e2eprechigh] {\prehoc};
\addlegendentry{E2E}

\addplot+[error bars/.cd, y dir = both, y explicit] table[x=metric, y=relprec, y error minus=relpreclow, y error plus=relprechigh] {\prehoc};
\addlegendentry{REL}

\nextgroupplot[title={Posthoc Verification}, ylabel={}]
\addplot+[error bars/.cd, y dir = both, y explicit] table[x=metric, y=e2eprec, y error minus=e2epreclow, y error plus=e2eprechigh] {\posthoc};
\addlegendentry{E2E}

\addplot+[error bars/.cd, y dir = both, y explicit] table[x=metric, y=relprec, y error minus=relpreclow, y error plus=relprechigh] {\posthoc};
\addlegendentry{REL}

\addplot+[error bars/.cd, y dir = both, y explicit]  table[x=metric, y=gtprec, y error minus=gtpreclow, y error plus=gtprechigh] {\posthoc};
\addlegendentry{GT}

\nextgroupplot[title={}, ylabel={Recall}]
\addplot+[error bars/.cd, y dir = both, y explicit]  table[x=metric, y=e2erecall, y error minus=e2erecalllow, y error plus=e2erecallhigh] {\prehoc};

\addplot+[error bars/.cd, y dir = both, y explicit]  table[x=metric, y=relrecall, y error minus=e2erecalllow, y error plus=e2erecallhigh] {\prehoc};

\nextgroupplot[title={}, ylabel={}]
\addplot+[error bars/.cd, y dir = both, y explicit]  table[x=metric, y=e2erecall, y error minus=e2erecalllow, y error plus=e2erecallhigh] {\posthoc};

\addplot+[error bars/.cd, y dir = both, y explicit]  table[x=metric, y=relrecall, y error minus=relrecalllow, y error plus=relrecallhigh] {\posthoc};

\addplot+[error bars/.cd, y dir = both, y explicit] table[x=metric, y=gtrecall, y error minus=gtrecalllow, y error plus=gtrecallhigh] {\posthoc};




\end{groupplot}
\end{tikzpicture}
    \caption{Precision and recall results from pre-annotation evaluation compared with the posthoc verification evaluation. Error bars represent 95\% confidence intervals on bootstrapped samples of the data. Posthoc verification (on right) returns substantially higher scores than the pre-annotation evalauation. The ground truth (GT) data was often verified at a lower rate than EL models.}
    \label{fig:prehoc}
\end{figure}

First, we re-tested the E2E and REL models and evaluated their micro precision and recall under the typical pre-annotation evaluation regime. These results are illustrated in Fig~\ref{fig:prehoc} (left). These results are nearly identical to those reported by related works~\cite{kolitsas2018end, van2020rel}. In our opinion, these state-of-the-art models provide meager performance under the standard evaluation regime.

\subsection{Posthoc Verification Evaluation} 

Our next task is to define appropriate evaluation metrics that can be used to compare the results of the posthoc verification experiment with results from the pre-annotation evaluation regime.

\paragraph{Verification Rate.} For each combination of dataset and model providing annotations, we compute the verification rate as the percentage of annotations that were verified. Formally, let $d\in\textrm{datasets}$; $m\in\textrm{models}$; and $V_{m,d}$ be the set of verified annotations in a pairing of $d$ and $m$ (a value of a cell in the right three columns of Tab~\ref{tab:basic_statistics}). Likewise, let $N_{d,m}$ be the pre-annotations of model $m$ on dataset $d$. We therefore define the verification rate of a dataset-model pair as $r_{m,d} = |V_{m,d}|/|N_{d,m}|$. Higher verification rates indicate that the dataset contains annotations and/or the model is capable of providing labels that pass human inspection.

\paragraph{Verification Union.} It is important to note that each model and dataset was evaluated by only a single worker. However, we were careful to assign each worker annotations randomly drawn from model/dataset combinations. This randomization largely eliminates biases in favor or against any model or dataset. Furthermore, this methodology provides for repetitions when annotations match exactly across models -- which is what models are optimized for in the first place! In this scenario the union of all non-exact, non-overlapping annotations provides a superset of annotations similar to how pooling is used in information retrieval evaluation to create a robust result set~\cite{zobel1998reliable}. Formally, we define the verification union of a dataset $d$ as $V_d=\bigcup_{m}{V_{m,d}}$.

\paragraph{Posthoc Precision and Recall.}
The precision metric is defined as the ratio of true predictions to all predictions. If we recast the concept of true predictions to be the set of verified annotations $V_{m,d}$, then it is natural to further consider $N_{d,m}$ to be the set of all predictions for some dataset and model pair, especially considering our data collection methodology restricts $V_{m,d}\subseteq N_{d,m}$. Thus the posthoc precision of a model-data pairing is simply the verification rate $r_{m,d}$.

The recall metric is defined as the ratio of true predictions to all true labels. If we keep the recasting of true positives as verified annotations $V_{m,d}$, then all that remains a definition of true labels. Like in most evaluation regimes the set of all true labels is estimated by the available labels in the dataset. Here, we do the same and estimate the set of true labels as the union of a dataset's verified annotations $V_d$. Thus posthoc recall of a model-data pairing is $|V_{m,d}|/|V_d|$.

\section{Posthoc Verification Results}

Using the evaluation tools introduced in the previous section, we begin to answer our original research questions. We begin with RQ1, which asks if the differences between evaluation regimes, \ie, pre-annotation versus posthoc verification, have any affect on our perception of model performance. To answer RQ1, we compared the precision and recall metrics calculated using the pre-annotation evaluation regime against the precision and recall metrics calculated using the posthoc verification regime.

Figure~\ref{fig:prehoc} compares model performance under the different evaluation regimes. Error bars represent the empirical 95\% confidence internals drawn from 1000 bootstrap samples of the data. We draw two major conclusions from this comparison: (1) Pre-annotation evaluation performance is much lower performance than Posthoc verification, and (2) the EL models usually outperform the ground truth, sometimes substantially.

\paragraph{Pre-annotation evaluation performance is lower than Posthoc verification.} The differences between the scores of the pre-annotation compared to posthoc verification are striking. Posthoc annotation shows nearly perfect precision scores across all datasets. Although the models may not exactly predict the pre-annotated label, high posthoc precision indicates that their results appear to be ``close-enough'' to pass a human's verification. 


In other words, the exact matching against pre-annotations is too strict. Despite its intention, the pre-annotation evaluation regime does not appear to faithfully simulate a human use case.

\begin{figure}
    \centering
    \pgfplotstableread{
model	e2everify	e2eedit	e2eremove	relverify	reledit	relremove	gtverify	gtedit	gtremove
AIDA-train	0.9135885008	0.07504596356	0.01136553568	0.8924074964	0.0902931283	0.0172993753	0.9019295256	0.08487713705	0.01319333736
AIDA-A	0.9099213937	0.07648183556	0.01359677077	0.887879911	0.09766493699	0.01445515196	0.8952139996	0.09487666034	0.00990934008
AIDA-B	0.9184149184	0.06783216783	0.01375291375	0.8952553277	0.08162444713	0.02312022517	0.9062993904	0.08421765636	0.009482953263
ACE2004	0.9178994083	0.06804733728	0.01405325444	0.8874849579	0.09927797834	0.01323706378	0.9598393574	0.03212851406	0.008032128514
AQUAINT	0.9296296296	0.05925925926	0.01111111111	0.9047619048	0.0642303433	0.03100775194	0.922425952	0.05359661495	0.023977433
CLUEWEB	0.9134235522	0.06833655543	0.01823989241	0.865775354	0.09780983376	0.03641481221	0.8809157542	0.09917654511	0.01990770066
MSNBC	0.9141494436	0.06995230525	0.01589825119	0.8950276243	0.0773480663	0.02762430939	0.9133126935	0.06656346749	0.02012383901
WIKIPEDIA	0.8983897141	0.08238671826	0.01922356759	0.8609337954	0.1181834762	0.02088272843	0.8795897131	0.09915266835	0.02125761855
}{\verify}

\edef\mylst{"An arbitrary string","String","Custom label","Not this data"}

\begin{tikzpicture}[/pgfplots/every axis/.append style={ 
ybar stacked,
/pgf/bar width=6pt,
legend columns=-1,
width = \linewidth, 
height = 1.5in,
ymin=0.8, ymax=1,
xlabel = {Dataset},
ylabel = {},
symbolic x coords={
    AIDA-train, 
    AIDA-A, 
    AIDA-B,
    ACE2004,
    AQUAINT,
    CLUEWEB,
    MSNBC,
    WIKIPEDIA,
},
xtick=data,
}
]

\begin{axis}[bar shift=-9pt, hide axis, 
legend style={font=\scriptsize,
	nodes={scale=1, transform shape},
	at={(0,1)},
	anchor=south west,
	draw=none, fill=none},] 
\addplot+[fill=white, draw=black, postaction={pattern=horizontal lines}] table[x=model,y=gtverify] {\verify};
\addlegendentry{E2E}
\addplot+[fill=white, draw=black, postaction={pattern=crosshatch}] table[x=model,y=gtedit] {\verify};
\addlegendentry{REL}
\addplot+[fill=white, draw=black, postaction={pattern=vertical lines}] table[x=model,y=gtremove] {\verify};
\addlegendentry{GT}
\end{axis}

\begin{axis}[bar shift=-9pt, hide axis, 
legend style={font=\scriptsize,
	nodes={scale=1, transform shape},
	at={(1,1)},
	anchor=south east,
	draw=none, fill=none},] 
\addplot+[] table[x=model,y=gtverify] {\verify};
\addlegendentry{Verify}
\addplot+[] table[x=model,y=gtedit] {\verify};
\addlegendentry{Edit}
\addplot+[] table[x=model,y=gtremove] {\verify};
\addlegendentry{Remove}
\end{axis}

\begin{axis}[bar shift=-9pt, hide axis] 
\addplot+[postaction={pattern=horizontal lines}] table[x=model,y=e2everify] {\verify};
\addplot+[postaction={pattern=horizontal lines}] table[x=model,y=e2eedit] {\verify};
\addplot+[postaction={pattern=horizontal lines}] table[x=model,y=e2eremove] {\verify};
\end{axis}

\begin{axis}[bar shift=0pt]
\addplot+[postaction={pattern=crosshatch}] table[x=model, y=relverify] {\verify};
\addplot+[postaction={pattern=crosshatch}] table[x=model,y=reledit] {\verify};
\addplot+[postaction={pattern=crosshatch}] table[x=model,y=relremove] {\verify};
\end{axis}

\begin{axis}[bar shift=9pt, hide axis] 
\addplot+[postaction={pattern=vertical lines}] table[x=model,y=gtverify] {\verify};
\addplot+[postaction={pattern=vertical lines}] table[x=model,y=gtedit] {\verify};
\addplot+[postaction={pattern=vertical lines}] table[x=model,y=gtremove] {\verify};
\end{axis}

\end{tikzpicture}
    \caption{Detailed error analysis of verification rates in Fig.~\ref{fig:prehoc}(top right). The E2E model consistently outperforms the ground truth dataset.
    }
    \label{fig:verifyrate}
\end{figure}
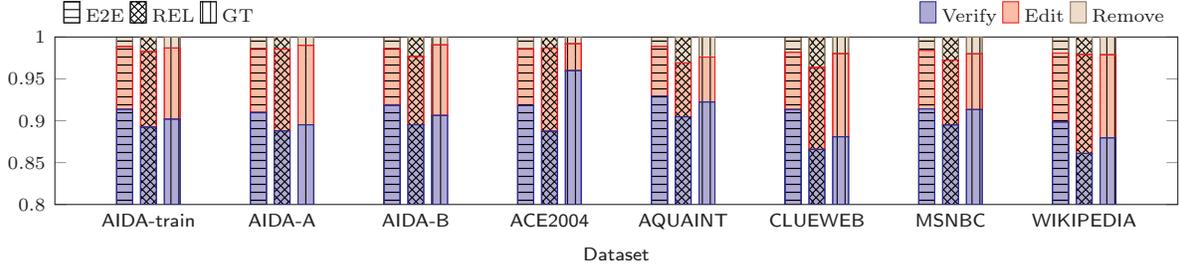

\paragraph{Machine Learning models outperform the Ground Truth. }
The posthoc verification methodology permits the ground truth annotations to be treated like any other model. So the right column in Fig.~\ref{fig:prehoc} includes the verification precision and recall for GT. The results were unexpected. We found that labels produced by the EL models oftentimes had a higher verification rate than the pre-annotated ground truth. The recall metric also showed that the EL models were also able to identify more verified labels than GT.

What does this mean?  Higher precision performance of the EL models indicates that human annotators make more unverifiable annotations than the EL models. Even the precision of AIDA\_train, which is trained \textit{and tested} on GT, has a higher verification rate than the GT (although not by much). Higher recall performance of the EL models indicates that the EL models find a greater coverage of possible entities. The recall results are less surprising because human annotators may be unmotivated or inattentive during free-form annotation -- qualities that tend to not affect EL models.

\paragraph{Error Analysis of the Ground Truth.} For each linked entity, the posthoc verification methodology permitted one of three outcomes: verification, modification, or removal. Figure~\ref{fig:verifyrate} shows the percentage of each outcome for each model and dataset pair; it is essentially a zoomed-in, more-detailed illustration of the top-right panel in Figure~\ref{fig:prehoc}, but with colors representing outcomes and patterns representing models.

Edits indicate that the named entity recognition (\ie, mention detection) portion of the EL model was able to identify an entity, but the entity was not linked to an verifiable entity. The available dataset has an enumeration of corrected linkages, but we do not consider them further in the present work. Removal indicates an error with the mention detection. From these results we find that, when a entity mention is detected it is usually a good detection; the majority of the error comes from the linking/disambiguation subtask.

A similar error analysis of missing entities is not permitted from the data collection methodology because we only ask workers to verify pre-annotated or predicted entities, not add missing entities. Because all detected mentions are provided with some entity link, we can safely assume that missing entities is mostly (perhaps wholly) due to errors in the mention detection portion of EL models.

\begin{figure}
\centering
    \pgfplotstableread{
model	v3   v2   v1 v0  
AIDA-train	0.8245187296	0.1385780805	0.03315032316	0.003752866773
AIDA-A	 0.826458037	0.1376955903	0.03214793741	0.003698435277
AIDA-B	 0.8450244698	0.1275693312	0.02349102773	0.003915171289
ACE2004     0.9440559441	0.04895104895	0.006993006993	0
AQUAINT     0.9017857143	0.08482142857	0.01339285714	0
CLUEWEB     0.8539109507	0.1135980746	0.02719614922	0.005294825511
MSNBC     0.8564356436	0.1113861386	0.01732673267	0.01485148515
WIKIPEDIA   0.8791064389	0.1024967148	0.01445466491	0.00394218134
}{\verify}

\begin{tikzpicture}
\sffamily
\begin{axis}[
ybar stacked,
bar width=6pt,
legend style={font=\scriptsize,
	nodes={scale=1, transform shape},
	at={(1,1)},
	anchor=south east,
	draw=none, fill=none},
legend columns=-1,
legend cell align={left},
width = .5\linewidth, 
height = 1.7in,
ymin=0.75, ymax=1,
ylabel = {Percent Agreement},
xlabel = {},
symbolic x coords={
    AIDA-train, 
    AIDA-A, 
    AIDA-B,
    ACE2004,
    AQUAINT,
    CLUEWEB,
    MSNBC,
    WIKIPEDIA,
},
xtick=data,
x tick label style={rotate=25,anchor=center, xshift=-0.4cm, yshift=-0.2cm},
]

\addplot+[] table[x=model, y=v3] {\verify};
\addlegendentry{$i=3$~~}

\addplot+[] table[x=model,y=v2] {\verify};
\addlegendentry{$i=2$~~}

\addplot+[] table[x=model,y=v1] {\verify};
\addlegendentry{$i=1$~~}

\addplot+[] table[x=model,y=v0] {\verify};
\addlegendentry{$i=0$~~}

\end{axis}
\end{tikzpicture}
    \label{fig:agreement}
    \caption{Agreement rates for three posthoc verification workers across an intersection of the results of all three models (E2E, REL, GT). $i=3$ indicates that all three workers verified the label, $i=2$ indicates that only two of the three verified the label, etc.}
    \label{fig:verify_agreement}
\end{figure}
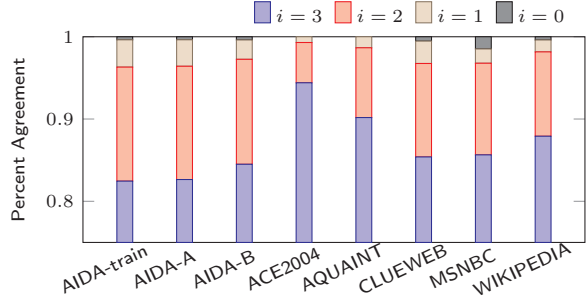

\subsection{Posthoc Verification Agreement}
Our second research question asks for an analysis of the agreement among the posthoc annotators. Just as agreement is assessed in pre-annotated ground truth data collection so as to guard against spurious or noisy data, so too is agreement useful in posthoc analysis. To answer RQ2, we first compute the intersection of all models for each dataset $I_d = \bigcap_{m}{N_{d,m}}$, that is, we only consider labels that were found in all three models. This severely restricted the analysis set in some datasets; in the most restrictive case, $I_{\scriptsize\textrm{ace2004}} = 246$ but its union contained 848. Nevertheless, from this reduced set we collected the labels $V^\prime_d = V_d\cap I_d$ that were verified by $i$ annotators.

Figure~\ref{fig:verify_agreement} shows that most annotations had complete agreement, \ie, all three annotators verified the same label for each of the three models. When one annotation was unverified, it still remained likely that the other annotators would still verify it. In less than 5\% of instances did two or more annotators decline to verify the label. These agreement rates are drastically higher than pre-annotation agreement rates, which had 16\% and 26\% agreements between three and two annotators respectively~\cite{botzer2021reddit}.

We also find that the ACE2004 and AQUAINT datasets appear to have a higher agreement rate than the other datasets. This begs the question: are the ACE2004 and AQUAINT datasets better benchmarks? Or is this simply a result of these low-recall datasets restricting the intersection to only the highest-quality (or easiest) annotations? This leads us to our final research question.

\subsection{Coverage of Verified Annotations in Ground Truth Datasets} 
We have previously seen that the pre-annotated ground truth datasets are slightly worse than some EL models. But what can these experiments tell us about RQ3? Can we evaluate the performance of the ground truth datasets relative to each other? Do some datasets contain more or fewer verifiable annotations? 

\begin{figure}
    \centering
    \pgfplotstableread{
metric	type_a type_b   type_c  type_d  type_e
AIDA-train	0.6811304938	0.2176360225	0.004630553671	0.002714462497	0.09388846753
AIDA-A	0.6523761081	0.2345589304	0.005377125418	0.004795814562	0.1028920215
AIDA-B	0.6461660079	0.2319367589	0.005059288538	0.006640316206	0.1101976285
ACE2004	0.2818396226	0.04245283019	0.01061320755	0.008254716981	0.6568396226
AQUAINT	0.4558599696	0.06925418569	0.02891933029	0.03729071537	0.4086757991
CLUEWEB	0.4730525736	0.1868978805	0.009193503991	0.01502890173	0.3158271401
MSNBC	0.6298342541	0.1668508287	0.02099447514	0.04751381215	0.1348066298
WIKIPEDIA	0.4283901784	0.1580714979	0.01888200612	0.04903350996	0.3456228077
}{\verifytype}

\begin{tikzpicture}
\sffamily

\begin{axis}[
xbar stacked,
/pgf/bar width=5pt,
legend style={font=\scriptsize,
	nodes={scale=1.0, transform shape},
	at={(1,1)},
	anchor=south east, 
	draw=none, fill=none},
legend columns=-1,
legend cell align={left},
width = .9\linewidth, 
height = 1.8in,
xmin=0.0, xmax=1,
xlabel = {Outcome},
ylabel = {Dataset},
symbolic y coords={
WIKIPEDIA,
MSNBC,
CLUEWEB,
AQUAINT,
ACE2004,
AIDA-B,
AIDA-A,
AIDA-train,
},
ytick=data,
]

\addplot+[] table[y=metric, x=type_a] {\verifytype};
\addlegendentry{A}

\addplot+[] table[y=metric, x=type_b] {\verifytype};
\addlegendentry{B}

\addplot+[] table[y=metric, x=type_c] {\verifytype};
\addlegendentry{C}

\addplot+[] table[y=metric, x=type_d] {\verifytype};
\addlegendentry{D}

\addplot+[] table[y=metric, x=type_e] {\verifytype};
\addlegendentry{E}

\end{axis}
\end{tikzpicture}
    \vspace{-.1cm}
    \caption{Outcomes from comparing union of verified annotations to GT annotations: (A) verified mention matches GT, verified link matches GT; (B) mention matches GT, link does not match GT; (C) mention overlaps another mention in GT, link matches GT; (D) mention overlaps GT, link does not match GT; (E) mention does not overlap with any mention in GT}
    \label{fig:five_agreements}
    \vspace{-.1cm}
\end{figure}
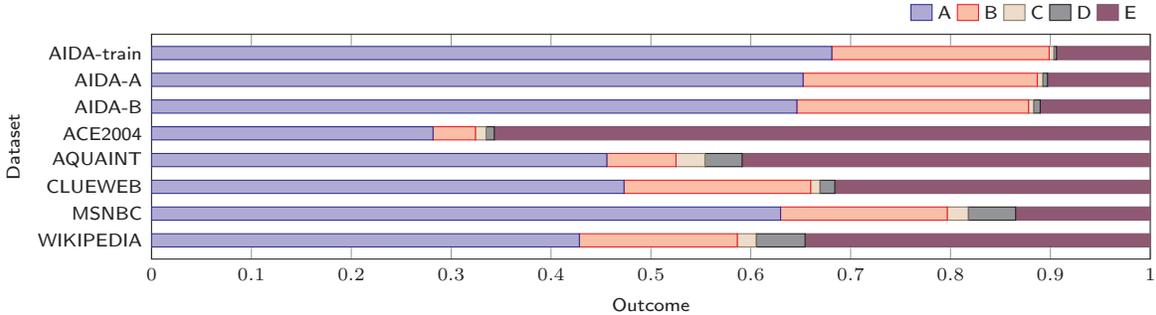

We answer this question by again considering, for each dataset, the union of verified annotations across all models $V_d$. Then, for each document in the dataset, we compare the annotations from $V_d$ to GT. For each annotation in $V_d$ this comparison can have five outcomes described in Fig.~\ref{fig:five_agreements}. Recall from Fig.~\ref{fig:prehoc}(top right) that many datasets had verification rates around 90\%. The ACE2004 dataset performed best with a verification rate of 96.4\%. But, we can observe from Fig.~\ref{fig:five_agreements} that this same dataset is missing nearly 65\% of verified annotations. Missing annotations (E) are generally the largest source of error in the ground truth, followed by mismatched--but still verified--entity links.

\section{Discussion and Recommendations}
\label{sec:discussion}
The primary goal of the present work is to compare pre-annotation labels contributed by human workers against verified annotations of the same data. Using entity linking as an example task, we ultimately found that these two methodologies returned vastly different performance results. From this observation we can draw several important conclusions. First, EL models have a much higher precision than related work reports. This difference is because the standard evaluation methodology used in EL, and throughout ML generally, do not account for soft matches or the semantics of what constitutes a label that is ``close enough''. Our second conclusion is that EL models, and perhaps ML models generally, sometimes perform better than ground truth annotators -- at least, that is, according to other ground truth annotators.

\paragraph{Limitations.}

The present work is not without its limitations. Here we only considered the entity linking task, which is only a small piece of ML research. We expect our results to generalize to any methodology that performs validation tests on subjective pre-annotation ground truth datasets, such as image labelling, 
social media analysis, text classification, and others. We also recognize that we used one set of annotators to argue against the work of, basically, the same set of annotators, \ie, turkers. Yet, our analysis is not meant to compare agreement among different sets of workers, but rather to show that the way in which we ask workers to perform their work is critical to the answer we receive. A final, rather obvious limitation of our methodology is cost. It costs money to ask workers to validate results. But in an era where millions of dollars are spent on training models, it seems reasonable to spend a few thousand on a more-thorough evaluation.

\paragraph{Recommendations.}
Based on the results of this paper, we recommend that ML researchers and practitioners reconsider the use of holdout test datasets based on pre-annotated data. In a time when improvement in machine learning models are measured by tenths of a percent, it is critical to remind ourselves that benchmark test sets, especially those comprised of pre-annotated labels are fallible -- and oftentimes less verifiable than the results of the models themselves. Our second recommendation is to reconsider the strict matching metrics for multiclass classification tasks like entity disambiguation and entity linking. We encourage the community to develop soft evaluation metrics like the BLEU score used in machine translation~\cite{papineni2002bleu} and the ROUGE score used in text summarization~\cite{lin2004rouge}.

\paragraph{Broader Impacts and Ethical Considerations.}
The present work is a call to re-evaluate the way that accuracy is evaluated in certain machine learning tasks. Implications of this work may redefine how researchers validate the state-of-the-art. We do not expect any negative broader impacts from this re-consideration of how machine learning models are evaluated. An impaneled ethical review board approved this experiment before data collection began. Human subjects were employed via Amazon Mechanical Turk to perform verification tasks. Tasks contained only pre-annotated data from existing datasets. Workers were provided an informed consent form, video, audio, and written instructions, as well as an example training task; no personal identifiable information (PII) was collected.



\bibliographystyle{abbrv}
\bibliography{refs}

\medskip

\small

\end{document}